\documentclass[letterpaper, 10 pt, conference]{ieeeconf} 
\usepackage{ulem}
\usepackage{multirow}
\usepackage{amsmath}
\usepackage{url}
\usepackage{cite}
\usepackage[font=footnotesize]{caption}
\usepackage{hyperref}
\usepackage{graphicx}
\usepackage{amssymb} 
\usepackage{tabularx}
\usepackage[table]{xcolor}
\usepackage{array}
\usepackage{subcaption}
\usepackage{booktabs}
\usepackage{makecell}

\newif\ifcolormode
\colormodefalse

\newcommand{\colortext}[1]{\ifcolormode\textcolor{blue}{#1}\else\textcolor{black}{#1}\fi}

\newcolumntype{M}[1]{>{\centering\arraybackslash}m{#1}}

\newcolumntype{?}{!{\vrule width 1pt}}

\usepackage{tikz}

\newcommand\submittedtext{%
  \footnotesize This work has been submitted to the IEEE for possible publication. Copyright may be transferred without notice, after which this version may no longer be accessible.}

\newcommand\submittednotice{%
\begin{tikzpicture}[remember picture,overlay]
\node[anchor=south,yshift=10pt] at (current page.south) {\fbox{\parbox{\dimexpr0.65\textwidth-\fboxsep-\fboxrule\relax}{\submittedtext}}};
\end{tikzpicture}%
}

\IEEEoverridecommandlockouts                              

\title{\LARGE \bf
SALSA: Swift Adaptive Lightweight Self-Attention \\ for Enhanced LiDAR Place Recognition
}

\author{Raktim Gautam Goswami$^{1}$, Naman Patel$^{1}$, Prashanth Krishnamurthy$^{1}$, Farshad Khorrami$^{1}$
\thanks{$^{1}$Control/Robotics Research Laboratory (CRRL), Department of Electrical and Computer Engineering, NYU Tandon School of Engineering, Brooklyn, NY, 11201. E-mails: {\tt\small \{rgg9769,nkp269, prashanth.krishnamurthy,khorrami\}@nyu.edu}
\newline
This work was supported in part by ARO under Grant W911NF-22-1-0028, and in part by the New York University Abu Dhabi (NYUAD) Center for Artificial Intelligence and Robotics (CAIR), funded by Tamkeen under the NYUAD Research Institute Award CG010.
}
}

\begin{document}




\maketitle
\thispagestyle{empty}
\pagestyle{empty}

\submittednotice

\begin{abstract}
Large-scale LiDAR mappings and localization leverage place recognition techniques to mitigate odometry drifts, ensuring accurate mapping. 
These techniques utilize scene representations from LiDAR point clouds to identify previously visited sites within a database. 
Local descriptors, assigned to each point within a point cloud, are aggregated to form a scene representation for the point cloud.
These descriptors are also used to re-rank the retrieved point clouds based on geometric fitness scores. 
We propose SALSA, a novel, lightweight, and efficient framework for LiDAR place recognition. It consists of a Sphereformer backbone that uses radial window attention to enable information aggregation for sparse distant points, an adaptive self-attention layer to pool local descriptors into tokens, and a multi-layer-perceptron Mixer layer for aggregating the tokens to generate a scene descriptor. The proposed framework outperforms existing methods on various LiDAR place recognition datasets in terms of both retrieval and metric localization while operating in real-time. 
The code is available at \href{https://github.com/raktimgg/SALSA}{https://github.com/raktimgg/SALSA}.  
\end{abstract}

\section{INTRODUCTION}
Place recognition plays a pivotal role in robot simultaneous localization and mapping (SLAM), enabling the retrieval of point clouds within a database that are near a query point cloud. Such capability is paramount for robot relocalization, where, given accurate data association, the query point cloud's pose can be deduced from retrieved point clouds. Relocalization and loop closure detection are essential for providing global constraints to mitigate drifts in SLAM.

The advent of deep learning techniques for modeling geometric information has markedly enhanced the relevance of LiDARs in place recognition for long-term SLAM. Additionally,  LiDARs have proven more robust compared to visual data in appearance-changing environments. Learning based LiDAR place recognition frameworks \cite{lcdnet,pointnetvlad,egonn,logg3dnet} transform geometric data (i.e., point clouds) into scene descriptor embeddings. They are trained so that embeddings corresponding to point clouds from nearby physical locations tend to be close in terms of Euclidean distance, while those from more distant locations are far apart. These embeddings are utilized for coarse localization by identifying the closest matches in the database for a given query point cloud. 
Subsequently, these matches are re-ranked based on geometric fitness scores for estimating the query point cloud's pose using registration.
\begin{figure}[htbp!]
\vspace*{-0.25cm}
\centering
\includegraphics[width=0.42\textwidth, trim={55 0 0 0}, clip]{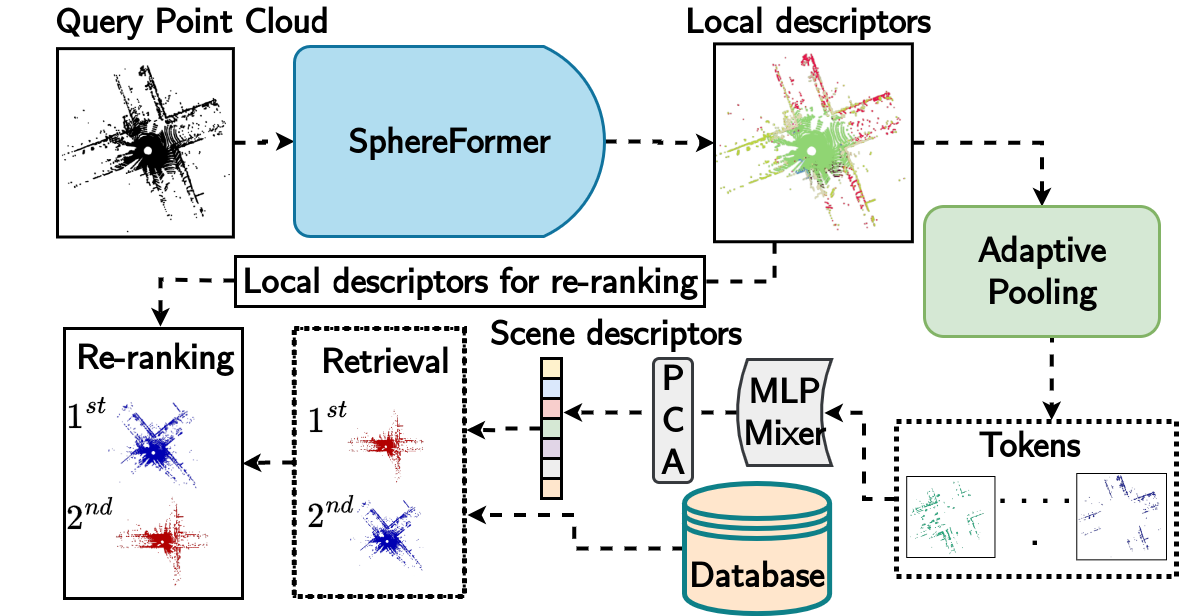}
    \caption{SALSA framework to retrieve and re-rank nearest point clouds using scene and local descriptors, respectively.}
    \label{fig:flow}
\end{figure}

Traditional LiDAR place recognition (LPR) systems use hand-crafted statistics and histogram-based approaches~\cite{classical4,scancontext,unlu2024efficient,bvmatch}, which make them unsuitable for describing large and complex scenes. With the introduction of large-scale datasets~\cite{alita,mulran,kitti,kitti360}, several deep learning based solutions have been proposed~\cite{pointnetvlad,egonn,lcdnet,logg3dnet}. These methods typically consist of a deep learning network that extracts descriptors for each point in a point cloud and an aggregator module to combine these features into a unified embedding. Despite significant advancements, existing state-of-the-art methods struggle to achieve accurate localization while being computationally efficient and lightweight. Thus, we propose SALSA which improves localization and retrieval performance on six large-scale datasets while being memory and compute efficient.

\begin{figure*}
\vspace*{5pt}
    \centering
    \includegraphics[width=0.8\textwidth, trim={0 0 10 0}, clip]{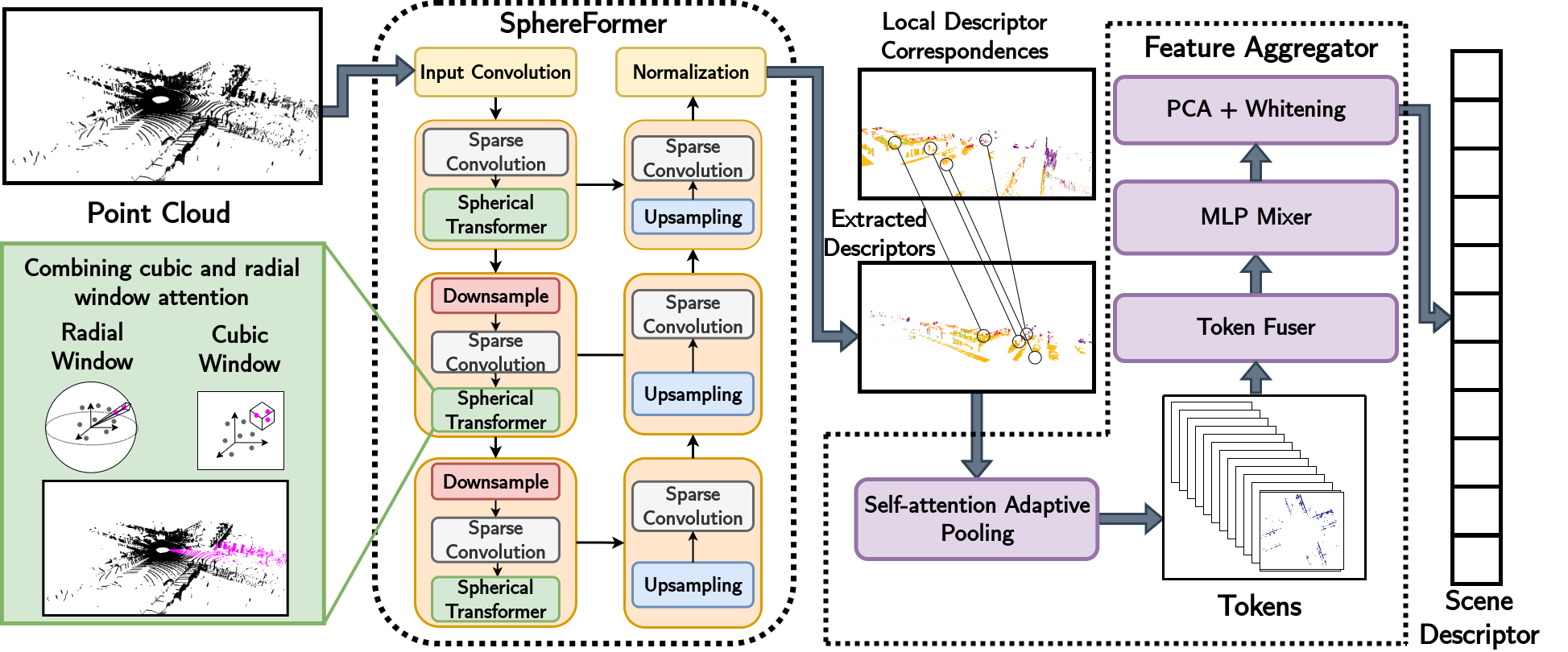}
    \caption{Overview of our SALSA framework to generate scene descriptors from point clouds for place recognition. A \textbf{SphereFormer} backbone with \textit{radial} and \textit{cubic} window attention is employed to extract local descriptors from point clouds. These local descriptors are fused into tokens via a \textbf{self-attention adaptive pooling} module. Subsequently, the pooled tokens are processed by an \textbf{MLP mixer}-based \textbf{aggregator} to iteratively incorporate global context information. 
    Finally, a \textit{PCA whitening} layer reduces the dimension and decorrelates the aggregated descriptor, producing the global scene descriptor.}
    \label{fig:architecture}
    \vspace*{-0.6cm}
\end{figure*}

We propose \textit{SALSA} (Fig. \ref{fig:flow}): an adaptive attention-based efficient framework for LiDAR place recognition that leverages a radial-window attention based SphereFormer~\cite{sphereformer} backbone and a novel aggregation method consisting of adaptive attention pooling, token fuser, and MLP-Mixer.
The Sphereformer backbone combines traditional voxel-based cubic window attention with radial window attention to extract local point descriptors. The spherical parameterization of the attention window allows attention between sparse distant points and effectively mitigates information disconnection issues caused by fewer neighborhood points. 
\colortext{Although our SphereFormer-based backbone leverages prior work, SALSA is the first to use radial-window attention in  LiDAR place recognition.}
This is followed by a self-attention adaptive pooling layer, which aggregates varying numbers of point features for a point cloud to a fixed number of tokens. Each token acts as an expert, attending to various parts of the scene, making it easier for the subsequent aggregator layer to filter non-informative regions (e.g., vehicles, ground plane) for generating a robust scene descriptor. The aggregator is a multi-layer-perceptron (MLP) Mixer~\cite{mlpmixer} that aggregates information from these tokens into a scene descriptor embedding. The embedding is further reduced in dimension and decorrelated using principal component analysis (PCA) with whitening to generate a scene descriptor for the point cloud. This pipeline enables robust retrieval of nearby point clouds for accurate localization in real-time. We further improve the retrieval performance by re-ranking the retrieved point clouds using corresponding compatibility graph-based spectral matching~\cite{spectralgv}. 
In summary, 
our contributions are:
\begin{itemize}
   \item  A lightweight feature pooling layer that adaptively pools relevant local features using self-attention into tokens.
    \item MLP mixer-based aggregator that combines information from the pooled tokens into a robust scene descriptor.
    \item \colortext{Introduce radial window attention in a LiDAR place recognition framework, which expands the local receptive field to better aggregate information from sparse, distant points to enhance local descriptor robustness.}
    \item Comprehensive evaluation of \textit{SALSA} on large LiDAR place recognition benchmarks to demonstrate its state-of-the-art performance and computational efficiency. 
\end{itemize}


\section{RELATED WORK}
\label{sec:related-work}
Early deep learning LiDAR place recognition approaches were able to leverage diverse techniques due to the availability of large datasets, compute, and algorithmic advancements.
These systems use point clouds or birds-eye-view (BEV) projection images as input and can be broadly divided into global descriptor based retrieval and local feature matching based methods. 
A pioneering method combines PointNet~\cite{pointnet} and NetVLAD~\cite{netvlad} to generate global features from point clouds~\cite{pointnetvlad}. Inspired by handcrafted bag-of-words features~\cite{bvmatch}, and \colortext{histogram bin-based encoded descriptors using BEV images that incorporate maximum point height~\cite{scancontext}}, intensity weighting~\cite{weighted-scancontext}, and semantics~\cite{seed}, frameworks employing group convolution on BEV images~\cite{bevplace} were developed to learn rotation-invariant local features.

Hybrid point-based approaches model point clouds' topological relationships, temporal consistencies, and structural appearance using 3D-CNNs~\cite{locus} or \colortext{attention graph neural networks on pose graphs, comparing nodes between sequential and non-sequential sub-graphs~\cite{pose-graph-lpr} for improved place recognition}. Similarly, adaptive local feature modules for handcrafted feature extraction are combined with graph-based neighborhood aggregation modules to discover the spatial distribution of local features in LPD-Net~\cite{lpdnet}. Enhanced local descriptors for 6-DoF pose refinement are obtained using 3D local feature encoders in DH3D~\cite{dhd}, while point orientation encoding (PointOE) modules are introduced in SOE-Net~\cite{soenet} to improve the distinctiveness of local features. \colortext{InCloud \cite{incloud} introduces an incremental learning approach to maintain the embedding space structure during fine-tuning on new scenes, addressing the challenge of learning from new domains without access to prior training data.} 

Transformer-based methods like TransLoc3D~\cite{transloc3d} and NDT-Transformer~\cite{ndttrans} utilize attention with adaptive receptive fields and normal distribution transform to enhance place recognition performance. \colortext{Spatial relationships at different scales are captured using a pyramid point transformer network in PPT-Net\cite{ppt-net}.}
However, MinkLoc3D~\cite{minkloc}, which employs sparse 3D convolutions, large batch training, and differentiable average precision loss~\cite{minklocv2} with fewer parameters, outperforms these transformer-based methods. Similarly, LCDNet~\cite{lcdnet}, LoGG3D-Net \cite{logg3dnet}, and EgoNN~\cite{egonn} also utilize sparse 3D convolutions for improved performance. LCDNet proposes differentiable relative pose heads for matching local features, while EgoNN uses keypoint detection and saliency prediction as proxies to generalize to large-scale scenes. LoGG3D-Net further improves performance by employing quadruplet loss with local consistency. To further boost the efficiency of LPR systems, re-ranking top retrieved matches is explored. SpectralGV~\cite{spectralgv} uses correspondence compatibility graphs from local descriptors to re-rank in sublinear time compared to time-consuming RANSAC~\cite{ransac} based geometric verification methods.

In contrast, SALSA employs the Sphereformer with radial window attention for sparse distant point aggregation alongside a novel pooling layer that adaptively projects features into tokens without downsampling, hence preserving information. Finally, the scene descriptor is generated by fusing these tokens using MLP mixer, which ensures faster computation for real-time place recognition, surpassing existing methods both with or without SpectralGV re-ranking.

\section{METHODOLOGY}
\label{sec:methodology}


\subsection{Problem Formulation}
\label{sec:problem-formulation}
The objective of our LiDAR place recognition framework is to learn an embedding function \(\Psi: \mathbb{R}^{N\times3} \rightarrow \mathbb{R}^{e}\) 
that maps a point cloud \(P_i \in \mathbb{R}^{N\times3}\) to an embedding. These embeddings should be closely aligned for point clouds that are geometrically close in terms of their 6-DoF LiDAR poses \(z_i\).
 \colortext{Thus, for a given query point cloud \(P_q\), its nearby positive point cloud \(P_p\), and farther away negative point cloud \(P_n\),}
\begin{align}
\label{eq:pf1}
    0&<d_E(z_q, z_p) \leq d_E(z_q, z_n) \nonumber\\
    &\implies 0<\|\Psi(P_q) - \Psi(P_p)\| \leq \|\Psi(P_q) - \Psi(P_n)\|
\end{align}
should hold, where \(d_E\) is the euclidean distance, \(P_i\) is a point cloud with pose \(z_i\)  for any $i$ with \(P_q, P_p, P_n\) being distinct. The function \(\Psi\) is expressed as a composite function with \(\Psi = g(h(.))\), with \(h(.): \mathbb{R}^{N\times 3} \rightarrow \mathbb{R}^{N\times d}\) serving as the feature extractor that assigns a local descriptor to each of the $N$ points within the point cloud and \(g(.):\mathbb{R}^{N\times d} \rightarrow \mathbb{R}^e\) as the aggregator of the descriptors.


\subsection{Feature Extraction}
\label{sec:backbone}
The SphereFormer~\cite{sphereformer} backbone is employed for point feature extraction, which leverages a U-Net~\cite{unet}. It combines sparse convolution-based decoder blocks with encoder blocks made of sparse convolutions and spherical transformers, as shown in Fig. \ref{fig:architecture}. The spherical transformer combines traditional cubic window attention with radial window attention for an increased receptive field for attention. Radial window attention partitions the space along the spherical coordinates $\alpha$ and $\beta$ axis with radius $r$ and assigns a window index, based on window size $\Delta \alpha$ and $\Delta \beta$, $win\_index_m = (\lfloor\frac{\alpha_m}{\Delta \alpha}\rfloor,\lfloor\frac{\beta_m}{\Delta\beta}\rfloor)$
to each point.
Points within the same partition are used to compute multi-headed attention. In contrast, cubic window attention methods categorize points based on their $x,y,z$ location within the cubic system. However, with LiDAR's inherent beam divergence, distant objects exhibit sparser geometry (e.g., spot size on the ground gets larger as distance increases), causing a drop in performance as cubic attention might have distant points in different windows. Radial window attention counters this by ensuring attention is performed for distant points within the same radial window. This results in superior feature descriptors for distant points, critical for feature matching, as depicted in Fig. \ref{fig:local-correcpondence}. Thus, to incorporate radial and local neighbor contextual information, both attention mechanisms are used by the encoder.
\begin{figure}[htbp!]
    \centering
\includegraphics[width=0.4\textwidth, trim={0 0 0 25}, clip]{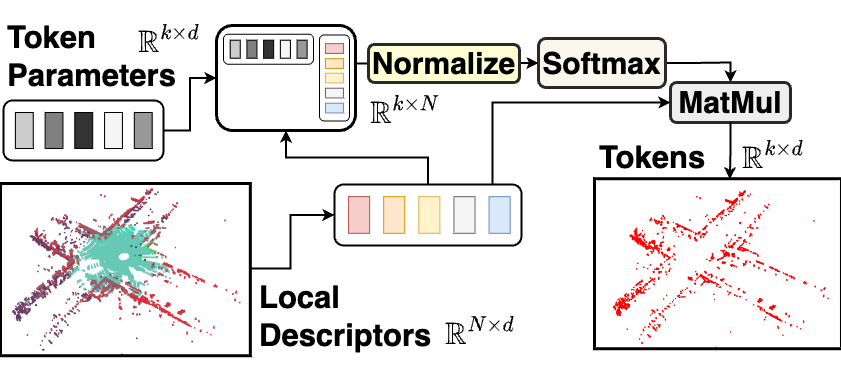}
    \caption{Our adaptive pooling module where local descriptors are pooled into tokens using self-attention mechanism.}
    \label{fig:attpool}
    \vspace*{-0.55cm}
\end{figure}
\subsection{Adaptive Attention Pooling}
\label{sec:adappool}
We make our framework computationally efficient and lightweight to enable large batch size training by pooling varying and large numbers of extracted local features into $k$ fixed tokens. The layer learns to adaptively pool features into tokens using the attention mechanism (Fig. \ref{fig:attpool}).  
The adaptive pooling layer uses an attention mechanism with trainable query parameters $Q_\theta \in \mathbb{R}^{k\times d}$ while utilizing the local features ($K\in \mathbb{R}^{N\times d}$) as both key and value:
\begin{align}
    A = \frac{\sigma(Q_\theta K^T)}{\sqrt{d}} \ \ ; \ \ 
    F = A K,
\end{align}
where $A$ contains the attention scores, $F$ $\in \mathbb{R}^{k \times d}$ are the tokens, and $\sigma$ is the softmax function applied along the rows. 
This layer learns to map uninformative geometries for localization tasks like moving cars, pedestrians, and roads to tokens separate from informative structures like intersections and trees, as shown in Fig. \ref{fig:attention-points}. This makes it easier for the token aggregator to generate a robust scene descriptor. 

Unlike prior methods that pool a fixed-number of descriptors, our approach pools a variable number of local descriptors using a learnable lookup query matrix ($Q_{\theta}$) for attention instead of quadratic self-attention operation. This is crucial for computational efficiency due to the varying number of points in each LiDAR scan.

\subsection{Token Aggregator}
\label{sec:aggregator}
We aggregate tokens $F$ from the pooling layer using a token aggregator that includes a token fuser and MLP Mixer~\cite{mlpmixer}. In the token fuser, tokens are iteratively normalized and processed through four 2-layer MLP blocks with ReLU and block residual connections, generating $k$ fused tokens of $d$ channels, where $k=512$ and $d=16$. This method enhances token information sharing~\cite{mixvpr} and improves speed and performance compared to attention.
\begin{figure}[!bthp]
    \centering
\includegraphics[width=0.4\textwidth, trim={0 80 0 25}, clip]{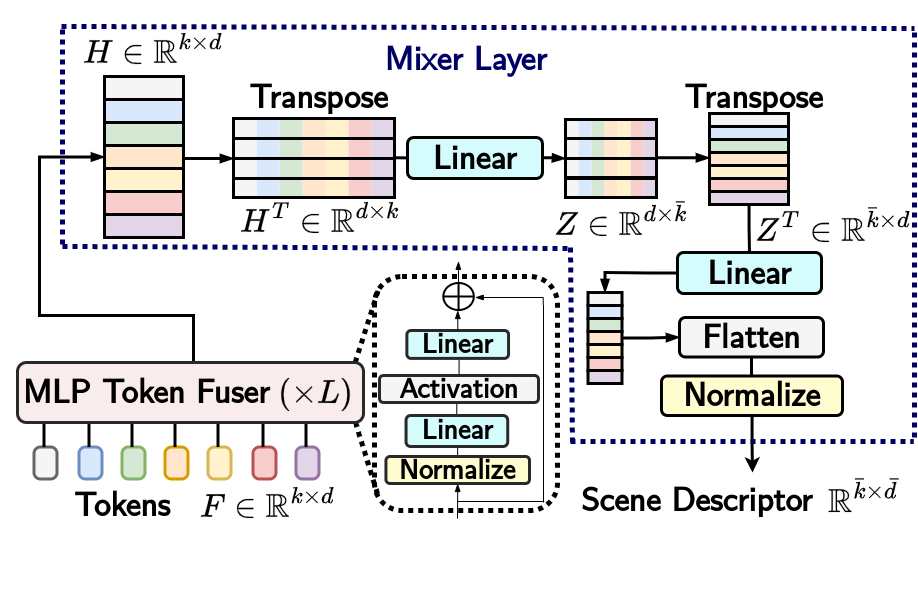}
    \caption{Our token aggregator, that iteratively fuses token embeddings to pool them into a scene descriptor using channel and token mixing MLPs.}
    \label{fig:mixer}
    \vspace*{-0.3cm}
\end{figure}
As depicted in Fig.~\ref{fig:mixer}, these fused tokens ($H \in \mathbb{R}^{k \times d}$) pass through channel-mixing and token-mixing MLPs for channel and token information fusion, respectively. The channel-mixing MLPs project the tokens to $\mathbb{R}^{\bar{k} \times d}$ along its row and token-mixing MLPs to $\mathbb{R}^{\bar{k} \times \bar{d}}$, along its column, with $\bar{d}=4$ and $\bar{k}=128$. The output is then flattened, normalized, and post-training, PCA with whitening is applied to reduce dimensionality and decorrelate the vectors.

\subsection{Re-ranking}
\label{sec:reranking}
SpectralGV~\cite{spectralgv} is used to re-rank the top retrieved candidates. It does not require registration of the point clouds and runs in sub-linear time compared to computationally inefficient RANSAC. It uses point correspondences to form a graph, where the edges between the corresponding points represent the spatial compatibility score. The score evaluates the retention of pairwise geometric relationships of each point pair in the query and retrieved point cloud. Points with high spatial compatibility are clustered together, and the maximum inter-cluster score is interpreted as the geometric fitness score to re-rank the retrieved candidates.

\subsection{Training }
\subsubsection{Partial hard negative mining}
\label{sec:mining}
We use partial hard negative mining~\cite{netvlad} for efficient training with triplets and help mitigate model collapse. A triplet consists of a query point cloud and its positive and negative point clouds, which are within a radius of 5m and outside 20m, respectively. 
The training dataset is divided into subsets, each with 1000 query point clouds. During training, for every subset, 4000 negatives are randomly sampled, and one positive is used for every query. A hard negative for a triplet is a point cloud whose descriptor distance is closer to the query than the positive's descriptor distance. We select the hardest negative for every query to generate the triplet and skip queries that don't have hard negatives for training.

\subsubsection{Training objective}
\label{sec:loss}
The training objective is a linear combination of a global triplet loss term for the point cloud and a local consistency loss term computed from the query and positive point clouds.
The triplet margin loss is given by 
\begin{align}
\resizebox{.9\hsize}{!}{$
    \mathcal{L}_G(q_i,p_i,n_i) = \max \{\|q_i-p_i\|_2^2 - \|q_i-n_i\|_2^2 + m,0\}, 
    $}
\end{align}
where $q_i, p_i, n_i$ are the embeddings of the query, positive and negative, respectively, with margin $m$ being 0.1.

Similar to \cite{logg3dnet}, the local consistency term employs a contrastive loss \cite{hardest-contrastive} to minimize distances between corresponding point pairs, while maximizing it for non-corresponding pairs in subset $\mathcal{Q}$. Defined using local features ${ {g}_1, {g}_2}$ and correspondences $\Gamma^{1\leftrightarrow2}$, the local consistency loss is:
\begin{align}
    &\mathcal{L}_l = \sum_{(i,j)\in \Gamma^{1\leftrightarrow2}}
    \{ [\|\{g_i\}_1-\{g_j\}_2\|_2^2-m_p]_+/|\Gamma^{1\leftrightarrow2}| \nonumber \\
    & + \mu_n I_i\left[ m_n - \min_{k\in\mathcal{Q}}\|\{g_i\}_1-\{g_k\}_2\|_2^2\right]_+/|\Gamma^{1\leftrightarrow2}_1| \nonumber\\
    & + \mu_n I_j\left[ m_n - \min_{k\in\mathcal{Q}}\|\{g_j\}_1-\{g_k\}_2\|_2^2\right]_+/|\Gamma^{1\leftrightarrow2}_2|\} ,
\end{align}
where $[]_+$ is the hinge loss, $\{g_i\}_1$ and $\{g_j\}_2$ are the descriptors of $i^{th}$ point of query and $j^{th}$ point of positive, respectively. $I_i = I(i,k_i,r)$ is an indicator function which is 1 if $k_i$ ($=argmin_{k \in \mathcal{Q}}\|g_i - g_k \|_2$) is located outside a radius of $r$ from point $i$, and 0 otherwise.  $|\Gamma^{1\leftrightarrow2}_1|$ is the number of valid mined negatives for points in the first point cloud, and $|\Gamma^{1\leftrightarrow2}_2|$ represents the same for the second point cloud.  $m_p, m_n, \mu_n$ are selected to be 0.1, 2, and 1, respectively.

\section{EXPERIMENTS}
\label{sec:experiments}
\subsection{Datasets and Evaluation Methodology}
\label{sec:datasets}
\begin{table*}
    \vspace*{5pt}
    \caption{Place recognition \colortext{(without re-ranking)} on the `easy' test setting.  R1: Recall@1; R5: Recall@5}
    \centering
    \setlength{\tabcolsep}{3pt}

    \begin{tabular}{? l ? c  c  c  c | c  c  c  c | c  c  c  c | c  c  c  c ?}
    \Xhline{3\arrayrulewidth}
     & \multicolumn{4}{c|}{\textbf{Mulran Sejong}} & \multicolumn{4}{c|}{\textbf{Apollo-Southbay}} & \multicolumn{4}{c|}{\textbf{KITTI}} & \multicolumn{4}{c?}{\textbf{Average}} \\
     \textbf{Methods} & \multicolumn{2}{c}{\textbf{5m}} & \multicolumn{2}{c|}{\textbf{20m}} & \multicolumn{2}{c}{\textbf{5m}} & \multicolumn{2}{c|}{\textbf{20m}} & \multicolumn{2}{c}{\textbf{5m}} & \multicolumn{2}{c|}{\textbf{20m}} & \multicolumn{2}{c}{\textbf{5m}} & \multicolumn{2}{c?}{\textbf{20m}} \\
     & \textbf{R1$(\uparrow)$} & \textbf{R5$(\uparrow)$} & \textbf{R1$(\uparrow)$} & \textbf{R5$(\uparrow)$} & \textbf{R1$(\uparrow)$} & \textbf{R5$(\uparrow)$} & \textbf{R1$(\uparrow)$} & \textbf{R5$(\uparrow)$} & \textbf{R1$(\uparrow)$} & \textbf{R5$(\uparrow)$} & \textbf{R1$(\uparrow)$} & \textbf{R5$(\uparrow)$} & \textbf{R1$(\uparrow)$} & \textbf{R5$(\uparrow)$} & \textbf{R1$(\uparrow)$} & \textbf{R5$(\uparrow)$} \\
    \Xhline{3\arrayrulewidth}
    
    MinkLoc3D \cite{minkloc} & 82.3 & 94.4 & 92.1 & 96.6 & 77.2 & 93.8 & 95 & 98.3 & 95.7 & 96.9 & 97.7 & 98 & 85.1 & 95.0 & 94.9 & 97.6 \\
    DiSCO \cite{disco} & 94 & 97.5 & 95.8 & 98.1 & 95.1 & 96.6 & 95.4 & 97 & 90.7 & 91.3 & 92.3 & 94.5 & 93.3 & 95.1 & 94.5 & 96.5 \\
    DH3D \cite{dhd} & 32.4 & 56.2 & 58.9 & 75.5 & 25.3 & 49.6 & 50.4 & 70.7 & 75.5 & 91.1 & 86.8 & 96.1 & 44.4 & 65.6 & 65.4 & 80.8 \\
    Locus \cite{locus} & 67 & 75.8 & 70.6 & 74.6 & 85.5 & 95.5 & 94.9 & 98.1 & 99 & {99.7} & 99.8 & 100 & 83.8 & 90.3 & 88.4 & 90.9 \\
    \colortext{PPT-Net} \cite{ppt-net} & 60.2 & 76.2 & 79.8 & 84.5 & 72.9 & 87.5 & 88.8 & 93 & 91 & 93.1 & 92.9 & 94.7 & 74.7 & 85.6 & 87.2 & 90.7 \\
    \colortext{MinkLoc3Dv2} \cite{minklocv2} & 73.1 & 86.4 & 87.1 & 90.9 & 89.1 & 96.6 & 97.9 & 99.1 & 92.9 & 94 & 95.6 & 96.9 & 85 & 92.3 & 93.5 & 95.6 \\
    EgoNN \cite{egonn} & \cellcolor{gray!25}\textbf{98.3} & \cellcolor{gray!25}\textbf{99.9} & \cellcolor{gray!25}\textbf{99.6} & \cellcolor{gray!25}\textbf{99.9} & {95.7} & 97.7 & 96.3 & 98.2 & 97.4 & 98.2 & 97.9 & 98.7 & 97.1 & 98.6 & 97.9 & 98.9 \\
    LCDNet \cite{lcdnet} & 63.1 & 82 & 85.8 & 92.1 & 64.7 & 81.7 & 87.4 & 92.1 & 96.9 & 99.4 & 99.5 & 99.8 & 74.9 & 87.7 & 90.9 & 94.7 \\
    LoGG3D-Net \cite{logg3dnet} & 97.7 & {99.7} & 98.8 & {99.8} & 95.2 & {98.4} & {98.2} & {99.2} & \cellcolor{gray!25}\textbf{99.7} & \cellcolor{gray!25}\textbf{99.8} & \cellcolor{gray!25}\textbf{100} & \cellcolor{gray!25}\textbf{100} & 97.5 & \cellcolor{gray!25}\textbf{99.3} & 99.0 & 99.6 \\
    \textbf{SALSA (Ours)} & {98.2} & 99.2 & {99} & 99.5 & \cellcolor{gray!25}\textbf{96.2} & \cellcolor{gray!25}\textbf{98.6} & \cellcolor{gray!25}\textbf{99.5} & \cellcolor{gray!25}\textbf{99.8} & {99} & 99.5 & {99.8} & \cellcolor{gray!25}\textbf{100} & \cellcolor{gray!25}\textbf{97.8} & {99.1} & \cellcolor{gray!25}\textbf{99.4} & \cellcolor{gray!25}\textbf{99.8} \\
    \Xhline{2\arrayrulewidth}
    
    \end{tabular}
    \label{tab:retrieval-easy}

\end{table*}

\begin{table*}
    \caption{Place recognition \colortext{(without re-ranking)} on the `hard' test setting.  R1: Recall@1; MRR: Mean Reciprocal Rank}
    \centering
    \setlength{\tabcolsep}{1.0pt}
    \begin{tabular}{ ?l?c c c c|c c c c|c c c c|c c c c?}
    \Xhline{3\arrayrulewidth}
     & \multicolumn{4}{c|}{\textbf{Mulran DCC}} & \multicolumn{4}{c|}{\textbf{ALITA}} & \multicolumn{4}{c|}{\textbf{KITTI360}} & \multicolumn{4}{c?}{\textbf{Average}} \\
     \textbf{Methods}& \multicolumn{2}{c}{\textbf{5m}} & \multicolumn{2}{c|}{\textbf{20m}} & \multicolumn{2}{c}{\textbf{5m}} & \multicolumn{2}{c|}{\textbf{20m}} & \multicolumn{2}{c}{\textbf{5m}} & \multicolumn{2}{c|}{\textbf{20m}} & \multicolumn{2}{c}{\textbf{5m}} & \multicolumn{2}{c?}{\textbf{20m}} \\
     & \textbf{R1$(\uparrow)$} & \textbf{MRR$(\uparrow)$} & \textbf{R1$(\uparrow)$} & \textbf{MRR$(\uparrow)$} & \textbf{R1$(\uparrow)$} & \textbf{MRR$(\uparrow)$} & \textbf{R1$(\uparrow)$} & \textbf{MRR$(\uparrow)$} & \textbf{R1$(\uparrow)$} & \textbf{MRR$(\uparrow)$} & \textbf{R1$(\uparrow)$} & \textbf{MRR$(\uparrow)$} & \textbf{R1$(\uparrow)$} & \textbf{MRR$(\uparrow)$} & \textbf{R1$(\uparrow)$} & \textbf{MRR$(\uparrow)$} \\
    \Xhline{3\arrayrulewidth}
    Locus \cite{locus} & 46.3 & 55.6 & 71 & 77.5 & {95.2} & {97.6} & \cellcolor{gray!25}\textbf{100} & \cellcolor{gray!25}\textbf{100} & {92.7} & {95.1} & 97 & {98.1} & {78.07} & 82.77 & 89.33 & 91.87 \\
    \colortext{PPT-Net} \cite{ppt-net} & 47.9 & 60.2 & 80.8 & 85.3 & 71.4 & 81.7 & 90.5 & 93.7 & 64.6 & 70.8 & 73.1 & 77.5 & 61.30 & 70.90 & 81.47 & 85.50 \\
    \colortext{MinkLoc3Dv2} \cite{minklocv2} & 62.2 & 73.2 & 89.3 & 91 & 79.8 & 87.7 & 94 & 96.8 & 62.1 & 66.7 & 69.6 & 73.8 & 68.03 & 75.87 & 84.30 & 87.20 \\
    EgoNN \cite{egonn} & \cellcolor{gray!25}\textbf{67.1} & \cellcolor{gray!25}\textbf{77.2} & 90.2 & 92.3 & 79.8 & 87.3 & 91.7 & 94.9 & 86.9 & 90.1 & 88.2 & 91.4 & 77.93 & 84.87 & 90.03 & 92.87 \\
    LCDNet \cite{egonn} & 57.7 & 71.6 & {90.9} & \cellcolor{gray!25}\textbf{93.8} & 90.5 & 94.6 & 96.4 & 97.8 & 85.9 & 89.7 & 95.7 & 96.6 & 78.03 & {85.30} & {94.33} & {96.07} \\
    LoGG3D-Net \cite{logg3dnet} & {66.8} & {76.7} & \cellcolor{gray!25}\textbf{91.2} & {92.9} & 15.5 & 28 & 26.2 & 40 & \cellcolor{gray!25}\textbf{93.2} & \cellcolor{gray!25}\textbf{95.5} & {97.5} & {98.1} & 58.50 & 66.73 & 71.63 & 77.00 \\
    \textbf{SALSA (Ours)}& \cellcolor{gray!25}\textbf{67.1} & 76.3 & 90.2 & 91 & \cellcolor{gray!25}\textbf{100} & \cellcolor{gray!25}\textbf{100} & \cellcolor{gray!25}\textbf{100} & \cellcolor{gray!25}\textbf{100} & 91.2 & 94.5 & \cellcolor{gray!25}\textbf{97.7} & \cellcolor{gray!25}\textbf{98.5} & \cellcolor{gray!25}\textbf{86.10} & \cellcolor{gray!25}\textbf{90.27} & \cellcolor{gray!25}\textbf{95.97} & \cellcolor{gray!25}\textbf{96.50} \\
    \Xhline{2\arrayrulewidth}
    
    \end{tabular}
    \label{tab:retrieval-hard}
    \vspace*{-0.3cm}
\end{table*}

        
        
        
        

        
        
    
Our framework's capabilities are tested on six large-scale LiDAR localization benchmarks, namely, Daejeon Convention Center (DCC) and Sejong from Mulran~\cite{mulran}, Apollo-Southbay~\cite{southbay}, KITTI~\cite{kitti}, ALITA~\cite{alita}, and KITTI360~\cite{kitti360}. We use the same training and evaluation split as described in  SpectralGV~\cite{spectralgv}. The evaluation set consists of `easy' and `hard' sets where the `easy' set is similar to the EgoNN~\cite{egonn} test set and the `hard' set consists of unseen environments to test the framework's generalization capabilities. 

We assess a framework's performance using Recall@$k$ (for $k=1$ and $5$ candidates), mean reciprocal rank (MRR) calculated as $\frac{1}{|\Theta|}\sum_{i=1}^{|\Theta|} \frac{1}{rank_i}$ where $rank_i$ is the rank of the first correct retrieval, and localization metrics including average relative rotation error (RRE) and relative translation error (RTE) for successful candidates, and success rate (S). Success is defined as the estimated pose being within 2m and $5^{\circ}$ of the ground truth. All metrics are expressed as percentages.
\colortext{The global pose of the query pointcloud is derived from the retrieved candidate point cloud's known pose and the estimated relative pose, determined through RANSAC-based registration. The initial point correspondences are established based on local descriptor distances and pruned if the points are 0.5m apart or if their edge length ratio is over 0.8, with the smaller edge as the numerator. From this filtered set, three correspondences estimate the relative pose using the Kabsch algorithm, iterating until the confidence surpasses 0.999, up to 10,000 times. 
While using an iterative closest point algorithm on the filtered set can further improve performance, we opted to maintain consistency with prior work by not incorporating it.}
Candidate point clouds are retrieved based on their scene descriptor distance, with correct retrieval defined by distances of 5m and 20m from the query. 
\vspace{-0.1cm}
\subsection{Results}
\vspace{-0.1cm}
\subsubsection{Retrieval Results}
The retrieval performance on `easy' and `hard' sets \colortext{without re-ranking} is detailed in Tab.~\ref{tab:retrieval-easy} and \ref{tab:retrieval-hard}, respectively, using the same evaluation metrics as in~\cite{spectralgv} for consistency. We compare with Locus~\cite{locus}, DISCO~\cite{disco}, DH3D~\cite{disco}, MinkLoc3D~\cite{minkloc}, MinkLoc3Dv2~\cite{minklocv2}, PPT-Net~\cite{ppt-net}, EgoNN~\cite{egonn}, LCDNet~\cite{lcdnet}, and LoGG3D-Net~\cite{logg3dnet}.
These results demonstrate the efficacy and superior performance of SALSA for place recognition. Although LoGG3D-Net achieves a marginally higher Recall@5 on the `easy' set, SALSA outperforms it in all other metrics on average.
\begin{figure}[!ht]
    \centering
    \begin{subfigure}[b]{\linewidth} 
        \centering
        \includegraphics[width=0.85\linewidth, trim={0 10 0 0}, clip]{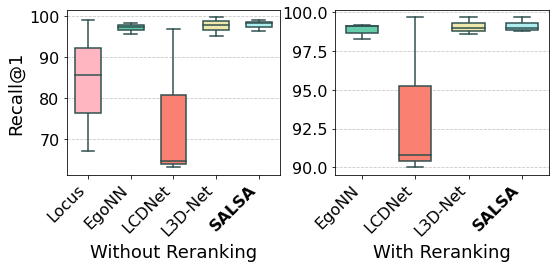}
        \caption{`Easy' Dataset} 
        \label{fig:box_plot_easy} 
    \end{subfigure}
    
    \par\medskip 

    \begin{subfigure}[b]{\linewidth} 
        \centering
        \includegraphics[width=0.85\linewidth, trim={0 10 0 0}, clip]{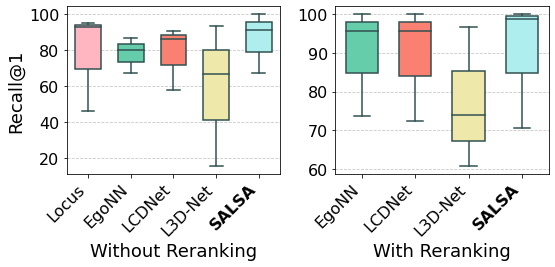}
        \caption{`Hard' Dataset} 
        \label{fig:box_plot_hard} 
    \end{subfigure}
    \caption{Box plot displaying Recall@1 across six datasets, with first to third quartile spans, whiskers for data variability, and internal lines as medians.}
    \vspace*{-0.7cm}
    \label{fig:model-spread} 
\end{figure} 

\subsubsection{Re-ranking Results}
The re-ranking performance evaluation of SALSA and other recent methods is shown in Tab.~\ref{tab:combined-table}. We apply the re-ranking method from SpectralGV~\cite{spectralgv} to the local descriptors of EgoNN~\cite{egonn}, LCDNet~\cite{lcdnet}, LoGG3D-Net~\cite{logg3dnet}, and SALSA (ours), re-ranking the top 20 candidates for each query as in SpectralGV~\cite{spectralgv}.

The spreads of the Recall@1 before and after re-ranking for best-performing models are plotted in Fig.~\ref{fig:model-spread}. 
The embedded lines inside each of the boxes signify the median. The boxes span from the first quartile $Q_1$ (median of the lower half of the data) to the third quartile $Q_3$  (median of the upper half). The whiskers above and below the boxes indicate the variability outside the interquartile range ($IQR = Q_3-Q_1$), i.e., the lower and higher whiskers are at $whis_1 = Q_1 - 1.5 IQR$ (or the smallest data point if greater than $whis_1$) and $whis_2 = Q_3 + 1.5 IQR$ (or the largest data point if smaller than $whis_2$), respectively.
SALSA maintains a high level of performance consistency, both with and without re-ranking, as shown in Fig. \ref{fig:model-spread}. The higher recall rates of SALSA in Tab.~\ref{tab:combined-table} further show its compatibility and enhanced performance with re-ranking methods like~\cite{spectralgv}.

\begin{table*}[ht]
\vspace{5pt}
    \centering
    \caption{Retrieval performance on `easy' and `hard' sets after re-ranking (left) in terms of R1 and R5 metrics. 6-DoF metric localization performance (right) with and without re-ranking in terms of success rate (S) and mean relative translation (RTE) and rotation (RRE) errors}
    \setlength{\tabcolsep}{0.25pt} 
    \begin{tabular}{?l ?c c c c ? c c c c ? c c c ? c  c c ?  c c  c ? c c c ?}
        \Xhline{3\arrayrulewidth}
        & \multicolumn{8}{c?}{\textbf{Place Recognition}} 
        & \multicolumn{12}{c?}{\textbf{6DoF Metric Localization}} \\

        & \multicolumn{8}{c?}{\textbf{Re-ranking using SpectralGV}} 
        & \multicolumn{6}{c}{\textbf{Without Re-ranking}} 
        & \multicolumn{6}{c?}{\textbf{Re-ranking using SpectralGV}} \\
        \Xhline{3\arrayrulewidth}
        & \multicolumn{4}{c?}{\textbf{Easy Datasets}} 
        & \multicolumn{4}{c?}{\textbf{Hard Datasets}} 
        & \multicolumn{3}{c?}{\textbf{Easy Datasets}} 
        & \multicolumn{3}{c?}{\textbf{Hard Datasets}} 
        & \multicolumn{3}{c?}{\textbf{Easy Dataset}} & \multicolumn{3}{c?}{\textbf{Hard Dataset}}\\
        
        \textbf{Methods} 
        & \textbf{R1(\(\uparrow\))} & \textbf{R5(\(\uparrow\))} & \textbf{R1(\(\uparrow\))} & \textbf{R5(\(\uparrow\))}
        & \textbf{R1(\(\uparrow\))} & \textbf{MRR(\(\uparrow\))} & \textbf{R1(\(\uparrow\))} & \textbf{MRR(\(\uparrow\))}
        & \textbf{S(\(\uparrow\))} & \textbf{RTE(\(\downarrow\))} & \textbf{RRE(\(\downarrow\))}
        & \textbf{S(\(\uparrow\))} & \textbf{RTE(\(\downarrow\))} & \textbf{RRE(\(\downarrow\))}
        & \textbf{S(\(\uparrow\))} & \textbf{RTE(\(\downarrow\))} & \textbf{RRE(\(\downarrow\))}  & \textbf{S(\(\uparrow\))} & \textbf{RTE(\(\downarrow\))} &\textbf{RRE(\(\downarrow\))}  \\
        \Xhline{2\arrayrulewidth}
        EgoNN& 98.8 & 99.3 & 99.4 & 99.6 & 89.77 & 93.03 & 97.93 & 98.17 & 97.7& 0.14& 0.37 & 87.6 & 0.14 & 0.52 & 98.9& 0.13& 0.37  & \cellcolor{gray!25}\textbf{94.6} & 0.13 & 0.53  \\
        LCDNet& 93.5 & 94.3 & 96.5 & 97.1 & 89.33 & \cellcolor{gray!25}\textbf{93.07} & \cellcolor{gray!25}\textbf{99.43} & \cellcolor{gray!25}\textbf{99.50} & 83.9& 0.19 & 0.45 & 87.5 & 0.16 & 0.53 & 95.2& 0.15 & 0.37  & 93.9 & 0.15 & 0.50  \\
        LoGG3D-Net& 99.1 & 99.6 & 99.7 & 99.8 & 77.10 & 81.10 & 90.20 & 91.10 & 98.9& 0.14 & 0.26 & 67.4 & 0.19 & 0.61 & \cellcolor{gray!25}\textbf{99.7}& 0.13 & 0.25  & 87.9 & 0.17 &0.51  \\
        \textbf{SALSA}& \cellcolor{gray!25}\textbf{99.2} & \cellcolor{gray!25}\textbf{99.6} & \cellcolor{gray!25}\textbf{99.8} & \cellcolor{gray!25}\textbf{99.9} & \cellcolor{gray!25}\textbf{89.80} & 92.53 & 96.10 & 96.53 & \cellcolor{gray!25}\textbf{99.1}& \cellcolor{gray!25}\textbf{0.13} & \cellcolor{gray!25}\textbf{0.25} &\cellcolor{gray!25}\textbf{92.0} & \cellcolor{gray!25}\textbf{0.11} & \cellcolor{gray!25}\textbf{0.33} & 99.6& \cellcolor{gray!25}\textbf{0.12} & \cellcolor{gray!25}\textbf{0.24}  & 93.0 & \cellcolor{gray!25}\textbf{0.10} & \cellcolor{gray!25}\textbf{0.31}  \\
        \Xhline{3\arrayrulewidth}
    \end{tabular}
    \label{tab:combined-table}
    \vspace*{-0.3cm}
\end{table*}

\begin{table*}[hbtp!]
    \centering
    \caption{Ablation studies on aggregator, pooling dimensions, and descriptor dimensions}
    \begin{subfigure}{0.32\linewidth}
        \centering
        \setlength{\tabcolsep}{1pt}
        \begin{tabular}{?c| c |c c c?}
            \Xhline{2\arrayrulewidth}
            \textbf{Backbone} & \textbf{Agg.} & \textbf{Params.} & \textbf{R1} & \textbf{MRR} \\
            \Xhline{2\arrayrulewidth}
            SPVCNN & LoGG3D& \cellcolor{gray!25}\textbf{9.45M} & 58.50 & 66.73\\
            SPVCNN (frozen) & SALSA& 9.6M& 60.10 & 67.90\\
            SPVCNN & SALSA& 9.6M & \cellcolor{gray!25}\textbf{83.93} & \cellcolor{gray!25}\textbf{88.73}\\
    
            \Xhline{1\arrayrulewidth}
            SALSA & SALSA &\cellcolor{gray!55}\textbf{3.74M} & \cellcolor{gray!55}\textbf{86.10} & \cellcolor{gray!55}\textbf{90.27}\\
            \Xhline{2\arrayrulewidth}
        \end{tabular}
        \caption{SALSA aggregator is integrated with SPVCNN (LoGG3D-Net backbone). The Recall@1 and the MRR scores are shown for the `hard' evaluation set}
        \label{tab:aggregator}
    \end{subfigure}\hspace{0.5cm}
    \begin{subfigure}{0.3\linewidth}
        \centering
        \setlength{\tabcolsep}{2pt}
        \begin{tabular}{?c | c |c c c?}
            \Xhline{2\arrayrulewidth}
            \textbf{Pool Dim.} & \textbf{Params.} & \textbf{D2} & \textbf{RV2} & \textbf{Avg.} \\
            \Xhline{2\arrayrulewidth}
            LoGG3D & 8.83M & 0.591 & 0.747 & 0.669 \\
            \Xhline{1\arrayrulewidth}
            256 & 3.59M & 0.652 & 0.947 & 0.799 \\
            512 & 3.74M & \cellcolor{gray!25}\textbf{0.679} & \cellcolor{gray!25}\textbf{0.981} & \cellcolor{gray!25}\textbf{0.830} \\
            1024 & 3.81M & 0.646 & 0.956 & 0.801 \\
            2048 & 3.96M & 0.639 & 0.955 & 0.803 \\
            \Xhline{2\arrayrulewidth}
        \end{tabular}
        \caption{Pooling Dimension: $F1_{max}$ evaluation scores on Mulran DCC2 (D2) and Riverside2 (RV2). Here, the descriptor dimension is kept constant at 512}
        \label{tab:model_performance_pool}
    \end{subfigure}\hspace{0.1cm}
    \begin{subfigure}{0.3\linewidth}
        \centering
        \setlength{\tabcolsep}{2pt}
        \begin{tabular}{?c | c | c c c?}
            \Xhline{2\arrayrulewidth}
            \textbf{Desc. Dim.} & \textbf{Params.} & \textbf{D2} & \textbf{RV2} & \textbf{Avg.} \\
            \Xhline{2\arrayrulewidth}
            LoGG3D & 8.83M & 0.591 & 0.747 & 0.669 \\
            \Xhline{1\arrayrulewidth}
            256& 3.60M& 0.633 & 0.970 & 0.801 \\
            512 & 3.74M & \cellcolor{gray!25}\textbf{0.679} & \cellcolor{gray!25}\textbf{0.981} & \cellcolor{gray!25}\textbf{0.830} \\
            1024 & 3.8M & 0.673 & 0.975 & 0.824 \\
            2048 & 3.94M & 0.674 & 0.972 & 0.823 \\
            \Xhline{2\arrayrulewidth}
        \end{tabular}
        \caption{Descriptor Dimension: $F1_{max}$ evaluation scores on Mulran DCC2 (D2) and Riverside2 (RV2). The pooling dimension is kept constant at 512}
        \label{tab:model_performance_descriptors}
    \end{subfigure}
    \label{tab:ablation}
    \vspace*{-0.5cm}
\end{table*}
\subsubsection{Metric Localization}
Results for 6-DoF metric localization on both test sets without and with re-ranking are shown in Tab.~\ref{tab:combined-table} with SALSA achieving high success rates and lowest average relative rotation and translation errors. 

\subsubsection{Discussion of Results}
\textit{Performance in `hard' set}: This set features shorter, unseen sequences with larger gaps between LiDAR scans, leading to a general performance decline across all methods. \textit{Tab.~\ref{tab:retrieval-easy} and \ref{tab:retrieval-hard}}: While LoGG3D-Net and EgoNN slightly outperform SALSA on KITTI and Mulran, respectively,  they underperform significantly on datasets like ALITA, where the data collection design markedly differs from the ones in the training data. Conversely, SALSA generalizes well to all datasets and obtains the best results on average, demonstrating its superior adaptiveness and resistance to overfitting. \textit{Post-re-ranking performance}: The performance of LCDNet after re-ranking increases by a higher percentage as compared to the other methods due to its large 640-dimension local descriptor, sacrificing inference speed (Fig. \ref{fig:runtime}). This necessitates downsampling points before aggregation, leading to mediocre outcomes in Tab.~ \ref{tab:retrieval-easy} and \ref{tab:retrieval-hard}.
In contrast, SALSA maintains high performance and speed with a smaller 16-dimensional descriptor, demonstrating robust local descriptors both before and after re-ranking.

\subsection{LiDAR SLAM and Loop Closure Detection}
\begin{figure}[!ht]
    \centering
    \begin{subfigure}[b]{0.42\linewidth} 
        \centering
        \includegraphics[width=\linewidth]{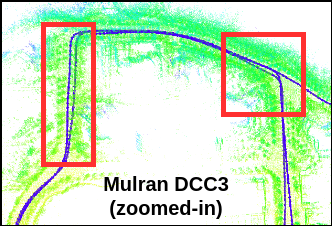}
        \caption{Without Loop Detection.} 
        \label{fig:loop-closure-a} 
    \end{subfigure}
    \begin{subfigure}[b]{0.42\linewidth} 
        \centering
        \includegraphics[width=\linewidth]{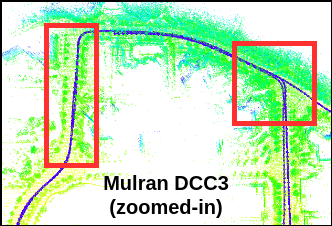}
        \caption{With Loop Detection.} 
        \label{fig:loop-closure-b} 
    \end{subfigure}
    \caption{Comparison of LiDAR-only odometry and maps: (a) without loop detection, and (b) after online pose graph optimization from SALSA loop detections. The highlighted rectangles emphasize the map and odometry disparities due to loop closures.}
    \vspace*{-0.8cm}
    \label{fig:loop-closure} 
\end{figure}

We show the map and odometry improvements by incorporating SALSA into a LiDAR-only SLAM system in Fig. \ref{fig:loop-closure}. We use KISS-ICP~\cite{kiss-icp} SLAM and compare its performance before and after applying online pose-graph optimization, utilizing the loop detection graph generated using SALSA. 

\subsection{Efficiency Analysis}
\begin{figure}[htbp!]
    \centering        \includegraphics[width=0.75\linewidth]{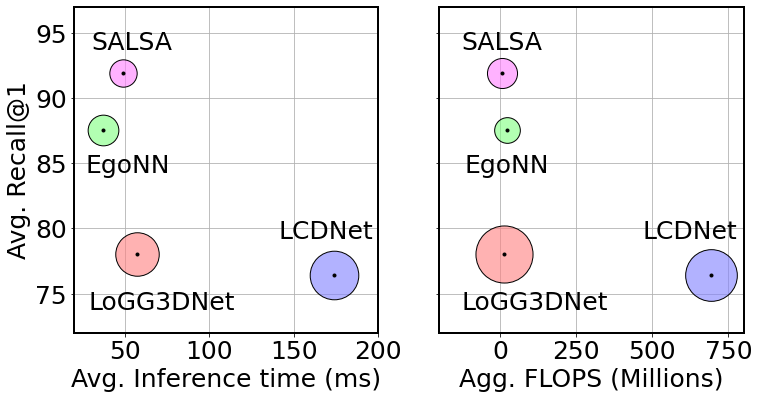}
        \caption{Runtime and memory analysis of place recognition models, with SALSA showing leading Recall@1 efficiency. 
        The left plot correlates Recall@1 to inference time, and the right to aggregator FLOPS, with circle sizes representing model parameters (left) and inference memory (right).}
        \vspace*{-0.4cm}
    \label{fig:runtime}
\end{figure}

The retrieval runtime and memory efficiency of SALSA (without re-ranking) are compared to other methods in Fig.~\ref{fig:runtime}. These frameworks are benchmarked on 621 query point clouds in KITTI Sequence 00. The average FLOPs are calculated for a point cloud with 65000 points. The left plot in Fig.~\ref{fig:runtime} shows Recall@1 performance against the inference time, and the right plot shows the number of FLOPS of their respective aggregators, respectively. The sizes of the circles in the figures represent the number of model parameters and its inference memory, respectively. \colortext{SALSA achieves an impressive inference time of just 0.049 seconds, which includes 0.017 seconds for point cloud pre-processing and 0.032 seconds for global descriptor calculation. When incorporating re-ranking, SALSA takes an additional 0.248 seconds, and the 6-DoF pose estimation requires 0.066 seconds.} It outperforms recent frameworks like EgoNN~\cite{egonn}, LoGG3D-Net~\cite{logg3dnet}, and LCDNet~\cite{lcdnet} in Recall@1 while running real-time with the lowest aggregator floating point operations (FLOPS) and lightest in terms of model parameters and inference memory. This demonstrate its ability to deliver high-performance place recognition while maintaining computational efficiency.

SALSA operates in real-time with an inference time of 49.4 ms, which is marginally slower than EgoNN. However, SALSA is 6$\%$ more accurate, giving the best performance while being the lightest with 3.74M parameters compared to 4.71M for EgoNN. \colortext{This superior performance is attributed to SALSA's ability to generate a high and variable number of local descriptors for each point cloud, outperforming EgoNN's fixed and smaller number of descriptors. SALSA can reduce 6-DoF pose estimation times to match that of EgoNN while maintaining superior performance by selecting fewer points with the highest attention scores from its adaptive attention pooling layer. This is demonstrated by utilizing 128 local point descriptors, similar to EgoNN, where SALSA achieves a 94.4$\%$ success rate, exceeding EgoNN's 92.65$\%$.}
\vspace{-0.5cm}
\vspace{-0.2cm}
\subsection{Qualitative Results}
\begin{figure}[htbp!]
    \centering
    \includegraphics[width = 0.6\linewidth]{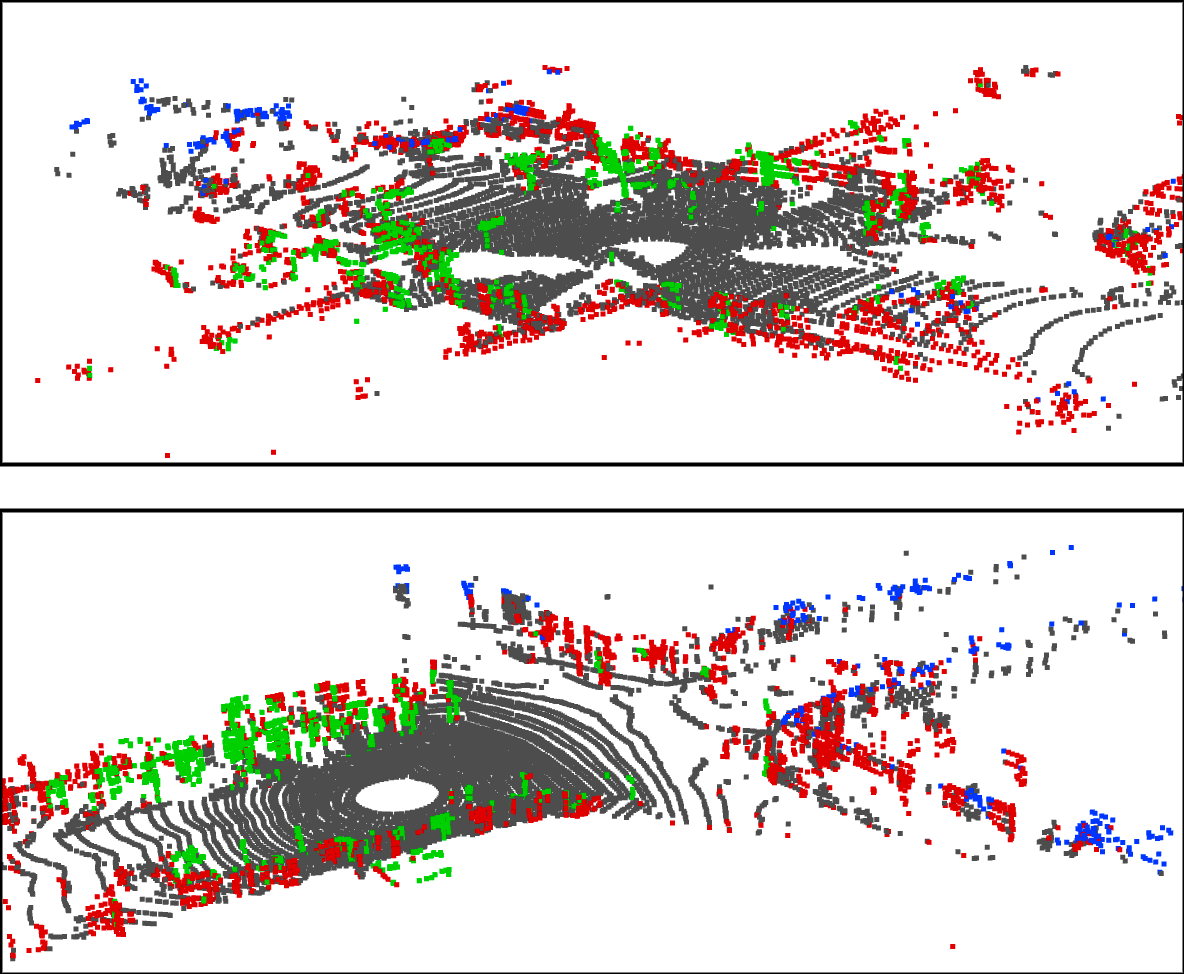}
    \caption{Visualization of areas attended to by different tokens from the adaptive pooling layer. Each token focuses on different geometries: trees and traffic signs (green), road intersections (red), and distant points (blue).}
    \vspace*{-0.5cm}
    \label{fig:attention-points}
\end{figure}

In Fig. \ref{fig:attention-points}, we show two examples that illustrate properties of three different tokens from the $k$ token output of the adaptive pooling layer (Sec. \ref{sec:adappool}) to selectively pass forward points from different geometries relevant to localization. Points with attention values above a threshold are plotted in green, blue, and red, while points below the threshold are gray. Green tokens attend to tall structures like trees and light posts, blue tokens focus on far-off points, and red tokens emphasize intersections. The attention mechanism effectively filters out non-informative points, such as the road.

\begin{figure}[htbp!]
    \centering
    \includegraphics[width = 0.77\linewidth, trim={0 0 0 0}, clip]{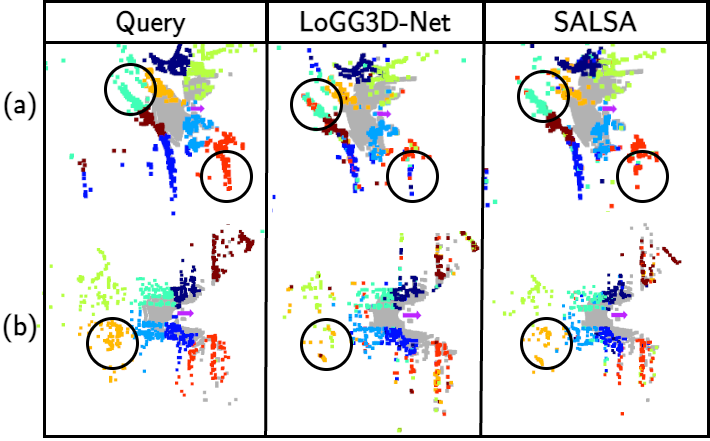}
    \caption{Point matches between query and target clouds using LoGG3D-Net and SALSA local descriptors. Matching colors indicate correspondences; circles highlight SALSA's superior performance on sparse distant points.}
     \vspace*{-0.5cm}
    \label{fig:local-correcpondence}
\end{figure}

Examples of local feature correspondences between query and target point clouds 5 meters apart are shown in Fig. \ref{fig:local-correcpondence} for LoGG3d-Net (one of the current best methods) and SALSA. Corresponding points are marked in the same color, with circles highlighting areas where SALSA shows correct correspondences while LoGG3D-Net fails. For instance, in the second example, LoGG3D-Net misses points and shows wrong correspondences (green points inside the circle), while SALSA captures correspondences with higher accuracy.
\subsection{Ablation Studies}
\subsubsection{Backbone}
\colortext{
The Sphereformer backbone used in SALSA is a U-Net~\cite{unet} with residual sparse 3D convolutions and a transformer block at each encoder block's end. The transformer block projects features into multiple heads, with half used by radial  and half by cubic window attention.}
\begin{table}[htbp!]
    \centering
    \caption{\colortext{Impact of various cubic and radial window attention configuration of SALSA backbone on retrieval performance, measured by Recall@1.}}
    \setlength{\tabcolsep}{3pt}
    \begin{tabular}{? c | c | c |c c | c ?} 
        \Xhline{2\arrayrulewidth}
        \textbf{Cubic} & \textbf{Radial} & \textbf{Params.} & \textbf{Easy Dataset} & \textbf{Hard Dataset} & \textbf{Average}\\
        \Xhline{2\arrayrulewidth}
        $\times$ & $\times$ & 3.40M & 95.3 & 83.6 & 89.45 \\
        \checkmark & $\times$ & 3.73M & 97.3 & 84.7 & 91.05 \\
        $\times$ & \checkmark & 3.73M & 96.9 & 84.0 &  90.45 \\
        \checkmark & \checkmark & 3.74M & \cellcolor{gray!25}\textbf{97.8} & \cellcolor{gray!25}\textbf{86.1} & \cellcolor{gray!25}\textbf{91.95}\\
        \Xhline{2\arrayrulewidth}
    \end{tabular}
  \label{tab:sphere-cube}
\end{table}
 \vspace*{-0.2cm}
 
\colortext{We evaluated these attention types on `easy' and `hard' LPR datasets as shown in Tab.~\ref{tab:sphere-cube}. Removing the Spherical Transformer blocks led to the worst performance. Cubic attention performed slightly better overall than radial but worse in some, as nearby dense objects are crucial for LPR with radial attention. Combining both attention yielded the most consistent and generalizable results across datasets.}

\subsubsection{Pooling and Embedding Dimension}
\label{sec:ablation_dim}
We found SALSA performed best when pooling and descriptor dimensions are set to 512.
It was evaluated based on $F1_{max}$ metric as shown in Tab.~\ref{tab:model_performance_pool} and Tab.~\ref{tab:model_performance_descriptors}, following the procedure of~\cite{logg3dnet} and also with ~\cite{logg3dnet} by the side. The model was trained on Mulran DCC1 and Riverside1, and tested on \colortext{DCC2 (D2) and Riverside2 (RV2)}. The pooling size is fixed at 512 while varying the descriptor dimensions and vice versa. 

\subsubsection{Aggregator}
We demonstrate the integration of SALSA's aggregator (adaptive pooling and MLP Mixer) with other LiDAR LPR backbone, SPVCNN from~\cite{logg3dnet} and frozen SPVCNN. The performance is evaluated based on Recall@1 and MRR scores on the `hard' evaluation set (Tab.~\ref{tab:aggregator}) with LoGG3D-net as baseline. Our aggregator performs better in both the cases with SPVCNN backbone than LoGG3D-net.

\begin{table}[!bhtp]
    \caption{Analysis of aggregator and pooling architectures: Recall@1 on Mulran DCC2 and Riverside2, Parameters, Memory (MB), and Time (ms).}
    \centering
    \setlength{\tabcolsep}{0.5pt}
    \begin{tabular}{?c | c | c c c | c c c?}
        \Xhline{2\arrayrulewidth}
        \textbf{Method} & \textbf{Params.} & \textbf{D2} & \textbf{RV2} & \textbf{Avg.} & \textbf{Agg. Param.} & \textbf{Mem.} & \textbf{Time} \\
        \Xhline{2\arrayrulewidth}
        SALSA & 3.74M & \cellcolor{gray!25}\textbf{0.69} & \cellcolor{gray!25}\textbf{0.99} & \cellcolor{gray!25}\textbf{0.85} & 0.208M & 50 & 1.1 \\
        \Xhline{1\arrayrulewidth}
        SALSA w/o token fuser & 3.73M & 0.66 & 0.99 & 0.83 & 0.205M & 50 & 0.7 \\
        $2^{nd}$ order pool & 3.55M & 0.63 & 0.98 & 0.81 & N/A & 98 & 2.0 \\
        NetVLAD & 3.89M & 0.64 & 0.95 & 0.79 & 0.339M & 50 & 0.9 \\
        GeM & 3.55M & 0.61 & 0.95 & 0.78 & N/A & 2 & 0.3 \\
        Global Avg. Pool (GP) & 3.42M & 0.55 & 0.78 & 0.67 & N/A & 2 & 0.3 \\
        \Xhline{1\arrayrulewidth}
        Adaptive Avg. Pool & 3.72M & 0.58 & 0.76 & 0.65 & - & - & - \\
        Random Pool & 3.72M & 0.01 & 0.04 & 0.03 & - & - & - \\
        \Xhline{2\arrayrulewidth}
    \end{tabular}
\label{tab:ablation_agg_pool}
\vspace{-0.5cm}
\end{table}
\begin{table}[!bhtp]
\caption{\colortext{Ablation study on different aggregators and backbones}}
\centering
\setlength{\tabcolsep}{2.5pt}
    \begin{tabular}{? c ? l  l  l ? l  l  l ? l  l  l ?}
    \Xhline{2\arrayrulewidth}
     \textbf{Backbones} $\rightarrow$ & \multicolumn{3}{c?}{\textbf{SALSA}} & \multicolumn{3}{c?}{\textbf{SPVCNN}} & \multicolumn{3}{c?}{\textbf{PTV3}} \\
    \Xhline{2\arrayrulewidth}
    \textbf{Aggregators} $\downarrow$ & \textbf{D2} & \textbf{RV2} & \textbf{Avg.} & \textbf{D2} & \textbf{RV2} & \textbf{Avg.} & \textbf{D2} & \textbf{RV2} & \textbf{Avg.} \\
    \Xhline{2\arrayrulewidth}
    SALSA & \cellcolor{gray!25}\textbf{0.69} & \cellcolor{gray!25}\textbf{0.99} & \cellcolor{gray!25}\textbf{0.84} & 0.5 & \cellcolor{gray!25}\textbf{0.92} & \cellcolor{gray!25}\textbf{0.71} & \cellcolor{gray!25}\textbf{0.65} & \cellcolor{gray!25}\textbf{0.99} & \cellcolor{gray!25}\textbf{0.82} \\
    $2^{nd}$ order pool & 0.63 & 0.98 & 0.81 & \cellcolor{gray!25}\textbf{0.59} & 0.75 & 0.67 & 0.16 & 0.21 & 0.19 \\
    NetVLAD & 0.64 & 0.95 & 0.79 & 0.52 & 0.79 & 0.65 & 0.64 & 0.89 & 0.77 \\
    GeM & 0.61 & 0.95 & 0.78 & 0.42 & 0.63 & 0.53 & 0.54 & 0.88 & 0.71 \\
    GP & 0.55 & 0.78 & 0.67 & 0.43 & 0.76 & 0.59 & 0.56 & 0.83 & 0.69 \\
    \Xhline{2\arrayrulewidth}
    \end{tabular}
    \label{tab:agg-back}
    \vspace{-0.2cm}
\end{table}
\colortext{Next, we compare SALSA's aggregator against others, focusing on $F1_{max}$ scores, parameter count, inference memory, and time (Tab. \ref{tab:ablation_agg_pool}). 
For all experiments, we used SALSA's backbone, training the models end-to-end with GeM, GP, aggregators and pooling methods, following the same training split as in Sec. \ref{sec:ablation_dim}.}
SALSA aggregator excels in performance and computational efficiency. 
Further, second-order pooling is memory-intensive, while GeM and GP have limited representational power. Random pooling is ineffective in encoding valuable information, and adaptive average pooling fails to generalize. The token fuser's significant role in SALSA is evident from the performance drop when it is removed.  
\colortext{
Additionally, to demonstrate the superior performance of SALSA's aggregator, we combine SALSA's aggregator and other popular aggregators (Tab. \ref{tab:agg-back}) with SPVCNN, the backbone of LoGG3D-Net \cite{logg3dnet} and PointTransformerV3 \cite{ptv3}. SALSA's aggregator consistently performs best with each backbone.}
\subsubsection{Random Rotation/Occlusion}
Evaluating SALSA with dynamic rotation of up to 30 degrees around the z-axis and removal of points in random 30-degree sectors yielded average $F1_{max}$ scores of 0.83 in both cases, demonstrating SALSA's robustness to rotation and occlusion.

\section{CONCLUSIONS}
\label{sec:conclusion}
We propose a lightweight, adaptive attention-based framework, SALSA, for real-time LiDAR place recognition. SALSA employs the Sphereformer backbone, which facilitates the fusion of features across sparse and distant points, while its adaptive pooling and aggregator modules enable memory-efficient information aggregation. The superior retrieval and localization performance of our efficient framework is demonstrated on various public benchmarks.






\bibliographystyle{IEEEtran}

\bibliography{IEEEabrv,IEEEexample}

\def\authornoop#1{}
\begin{thebibliography}{10}
\providecommand{\url}[1]{#1}
\csname url@rmstyle\endcsname
\providecommand{\newblock}{\relax}
\providecommand{\bibinfo}[2]{#2}
\providecommand\BIBentrySTDinterwordspacing{\spaceskip=0pt\relax}
\providecommand\BIBentryALTinterwordstretchfactor{4}
\providecommand\BIBentryALTinterwordspacing{\spaceskip=\fontdimen2\font plus
\BIBentryALTinterwordstretchfactor\fontdimen3\font minus \fontdimen4\font\relax}
\providecommand\BIBforeignlanguage[2]{{%
\expandafter\ifx\csname l@#1\endcsname\relax
\typeout{** WARNING: IEEEtran.bst: No hyphenation pattern has been}%
\typeout{** loaded for the language `#1'. Using the pattern for}%
\typeout{** the default language instead.}%
\else
\language=\csname l@#1\endcsname
\fi
#2}}

\bibitem{lcdnet}
D.~Cattaneo, M.~Vaghi, and A.~Valada, ``{LCDNet}: Deep loop closure detection and point cloud registration for lidar slam,'' \emph{IEEE Transactions on Robotics}, vol.~38, pp. 2074--2093, 2022.

\bibitem{pointnetvlad}
M.~A. Uy and G.~H. Lee, ``{PointNetVLAD}: Deep point cloud based retrieval for large-scale place recognition,'' in \emph{Proc. of the IEEE Conf. on Computer Vision and Pattern Recognition}, Salt Lake City, UT, June 2018, pp. 4470--4479.

\bibitem{egonn}
J.~Komorowski, M.~Wysoczanska, and T.~Trzcinski, ``{EgoNN}: Egocentric neural network for point cloud based 6dof relocalization at the city scale,'' \emph{IEEE Robotics and Automation Letters}, vol.~7, pp. 722--729, 2021.

\bibitem{logg3dnet}
K.~Vidanapathirana, M.~Ramezani, P.~Moghadam, S.~Sridharan, and C.~Fookes, ``{LoGG{3D}-Net}: Locally guided global descriptor learning for {{3D}} place recognition,'' in \emph{Proc. of the IEEE International Conf. on Robotics and Automation}, Philadelphia, PA, May 2022, pp. 2215--2221.

\bibitem{classical4}
L.~He, X.~Wang, and H.~Zhang, ``{M2DP}: A novel {3D} point cloud descriptor and its application in loop closure detection,'' in \emph{Proc. of the IEEE/RSJ International Conf. on Intelligent Robots and Systems}, Daejeon, South Korea, October 2016, pp. 231--237.

\bibitem{scancontext}
G.~Kim and A.~Kim, ``Scan context: Egocentric spatial descriptor for place recognition within {3D} point cloud map,'' in \emph{Proc. of the IEEE/RSJ International Conf. on Intelligent Robots and Systems}, Madrid, Spain, October 2018, pp. 4802--4809.

\bibitem{unlu2024efficient}
H.~U. Unlu, A.~Tzes, P.~Krishnamurthy, and F.~Khorrami, ``Efficient and distributed large-scale {3D} map registration using tomographic features,'' \emph{arXiv preprint arXiv:2406.19461}, 2024.

\bibitem{bvmatch}
L.~Luo, S.-Y. Cao, B.~Han, H.-L. Shen, and J.~Li, ``{BVMatch}: Lidar-based place recognition using bird's-eye view images,'' \emph{IEEE Robotics and Automation Letters}, vol.~6, pp. 6076--6083, 2021.

\bibitem{alita}
P.~Yin, S.~Zhao, R.~Ge, I.~Cisneros, R.~Fu, J.~Zhang, H.~Choset, and S.~Scherer, ``Alita: A large-scale incremental dataset for long-term autonomy,'' \emph{arXiv preprint arXiv:2205.10737}, 2022.

\bibitem{mulran}
G.~Kim, Y.~S. Park, Y.~Cho, J.~Jeong, and A.~Kim, ``Mulran: Multimodal range dataset for urban place recognition,'' in \emph{Proc. of the IEEE International Conf. on Robotics and Automation}, held virtually, May-August 2020, pp. 6246--6253.

\bibitem{kitti}
A.~Geiger, P.~Lenz, C.~Stiller, and R.~Urtasun, ``Vision meets robotics: The {KITTI} dataset,'' \emph{The International Journal of Robotics Research}, vol.~32, pp. 1231--1237, 2013.

\bibitem{kitti360}
Y.~Liao, J.~Xie, and A.~Geiger, ``{KITTI-360}: A novel dataset and benchmarks for urban scene understanding in {2D} and {3D},'' \emph{IEEE Transactions on Pattern Analysis and Machine Intelligence}, vol.~45, pp. 3292--3310, 2022.

\bibitem{sphereformer}
X.~Lai, Y.~Chen, F.~Lu, J.~Liu, and J.~Jia, ``Spherical transformer for lidar-based {3D} recognition,'' in \emph{Proc. of the IEEE Conf. on Computer Vision and Pattern Recognition}, Vancouver, Canada, June 2023, pp. 17\,545--17\,555.

\bibitem{mlpmixer}
I.~O. Tolstikhin, N.~Houlsby, A.~Kolesnikov, L.~Beyer, X.~Zhai, T.~Unterthiner, J.~Yung, A.~Steiner, D.~Keysers, J.~Uszkoreit, \emph{et~al.}, ``{MLP-Mixer}: An all-{MLP} architecture for vision,'' \emph{Advances in neural information processing systems}, vol.~34, pp. 24\,261--24\,272, 2021.

\bibitem{spectralgv}
K.~Vidanapathirana, P.~Moghadam, S.~Sridharan, and C.~Fookes, ``{Spectral Geometric Verification}: Re-ranking point cloud retrieval for metric localization,'' \emph{IEEE Robotics and Automation Letters}, vol.~8, pp. 2494--2501, 2023.

\bibitem{pointnet}
C.~R. Qi, H.~Su, K.~Mo, and L.~J. Guibas, ``{PointNet}: Deep learning on point sets for {3D} classification and segmentation,'' in \emph{Proc. of the IEEE Conf. on Computer Vision and Pattern Recognition}, Honolulu, HI, July 2017, pp. 652--660.

\bibitem{netvlad}
R.~Arandjelovic, P.~Gronat, A.~Torii, T.~Pajdla, and J.~Sivic, ``{NetVLAD}: {CNN} architecture for weakly supervised place recognition,'' in \emph{Proc. of the IEEE Conf. on Computer Vision and Pattern Recognition}, Las Vegas, NV, June 2016, pp. 5297--5307.

\bibitem{weighted-scancontext}
X.~Cai and W.~Yin, ``{Weighted Scan Context}: Global descriptor with sparse height feature for loop closure detection,'' in \emph{Proc. of the International Conf. on Computer, Control and Robotics}, Shanghai, China, January 2021, pp. 214--219.

\bibitem{seed}
Y.~Fan, Y.~He, and U.-X. Tan, ``Seed: A segmentation-based egocentric {3D} point cloud descriptor for loop closure detection,'' in \emph{Proc. of the IEEE/RSJ International Conf. on Intelligent Robots and Systems}, held virtually, October 2020, pp. 5158--5163.

\bibitem{bevplace}
L.~Luo, S.~Zheng, Y.~Li, Y.~Fan, B.~Yu, S.-Y. Cao, J.~Li, and H.-L. Shen, ``Bevplace: Learning lidar-based place recognition using bird's eye view images,'' in \emph{Proc. of the IEEE International Conf. on Computer Vision}, Paris, France, October 2023, pp. 8700--8709.

\bibitem{locus}
K.~Vidanapathirana, P.~Moghadam, B.~Harwood, M.~Zhao, S.~Sridharan, and C.~Fookes, ``Locus: Lidar-based place recognition using spatiotemporal higher-order pooling,'' in \emph{Proc. of the IEEE International Conf. on Robotics and Automation}, Xi'an, China, June 2021, pp. 5075--5081.

\bibitem{pose-graph-lpr}
M.~Ramezani, L.~Wang, J.~Knights, Z.~Li, P.~Pounds, and P.~Moghadam, ``Pose-graph attentional graph neural network for lidar place recognition,'' \emph{IEEE Robotics and Automation Letters}, vol.~9, pp. 1182--1189, 2023.

\bibitem{lpdnet}
Z.~Liu, S.~Zhou, C.~Suo, P.~Yin, W.~Chen, \emph{et~al.}, ``{LPD-Net}: {3D} point cloud learning for large-scale place recognition and environment analysis,'' in \emph{Proc. of the IEEE International Conf. on Computer Vision}, Seoul, Korea, October 2019, pp. 2831--2840.

\bibitem{dhd}
J.~Du, R.~Wang, and D.~Cremers, ``{DH{3D}}: Deep hierarchical {3D} descriptors for robust large-scale 6dof relocalization,'' in \emph{Proc. of the European Conf. on Computer Vision}, held virtually, August 2020, pp. 744--762.

\bibitem{soenet}
Y.~Xia, Y.~Xu, S.~Li, R.~Wang, J.~Du, D.~Cremers, and U.~Stilla, ``{SOE-Net}: A self-attention and orientation encoding network for point cloud based place recognition,'' in \emph{Proc. of the IEEE Conf. on Computer Vision and Pattern Recognition}, June 2021, pp. 11\,348--11\,357.

\bibitem{incloud}
J.~Knights, P.~Moghadam, M.~Ramezani, S.~Sridharan, and C.~Fookes, ``Incloud: Incremental learning for point cloud place recognition,'' in \emph{Proc. of the IEEE International Conf. on Robotics and Automation}, Kyoto, Japan, October 2022, pp. 8559--8566.

\bibitem{transloc3d}
T.-X. Xu, Y.-C. Guo, Z.~Li, G.~Yu, Y.-K. Lai, and S.-H. Zhang, ``{TransLoc{3D}}: point cloud based large-scale place recognition using adaptive receptive fields,'' \emph{Communications in Information and Systems}, vol.~23, pp. 57--83, 2023.

\bibitem{ndttrans}
Z.~Zhou, C.~Zhao, D.~Adolfsson, S.~Su, Y.~Gao, T.~Duckett, and L.~Sun, ``{NDT-Transformer}: Large-scale {3D} point cloud localisation using the normal distribution transform representation,'' in \emph{Proc. of the IEEE International Conf. on Robotics and Automation}, Xi'an, China, May 2021, pp. 5654--5660.

\bibitem{ppt-net}
L.~Hui, H.~Yang, M.~Cheng, J.~Xie, and J.~Yang, ``Pyramid point cloud transformer for large-scale place recognition,'' in \emph{Proc. of the IEEE/CVF International Conf. on Computer Vision}, held virtually, June 2021, pp. 6098--6107.

\bibitem{minkloc}
J.~Komorowski, ``{MinkLoc{3D}}: Point cloud based large-scale place recognition,'' in \emph{Proc. of the IEEE Winter Conf. on Applications of of Computer Vision}, held virtually, January 2021, pp. 1790--1799.

\bibitem{minklocv2}
J.~{\authornoop{T}}Komorowski, ``Improving point cloud based place recognition with ranking-based loss and large batch training,'' in \emph{Proc. of the International Conf. on Pattern Recognition}, Montréal Québec, Canada, August 2022, pp. 3699--3705.

\bibitem{ransac}
M.~A. Fischler and R.~C. Bolles, ``Random sample consensus: a paradigm for model fitting with applications to image analysis and automated cartography,'' \emph{Communications of the ACM}, vol.~24, pp. 381--395, 1981.

\bibitem{unet}
O.~Ronneberger, P.~Fischer, and T.~Brox, ``{U-Net}: Convolutional networks for biomedical image segmentation,'' in \emph{Proc. of the Medical Image Computing and Computer-Assisted Intervention}, Munich, Germany, October 2015, pp. 234--241.

\bibitem{mixvpr}
A.~Ali-Bey, B.~Chaib-Draa, and P.~Giguere, ``{MixVPR}: Feature mixing for visual place recognition,'' in \emph{Proc. of the IEEE Winter Conf. on Applications of of Computer Vision}, Waikoloa, HI, January 2023, pp. 2998--3007.

\bibitem{hardest-contrastive}
C.~Choy, J.~Park, and V.~Koltun, ``Fully convolutional geometric features,'' in \emph{Proc. of the IEEE International Conf. on Computer Vision}, Seoul,Korea, October 2019, pp. 8958--8966.

\bibitem{disco}
X.~Xu, H.~Yin, Z.~Chen, Y.~Li, Y.~Wang, and R.~Xiong, ``{DiSCO}: Differentiable scan context with orientation,'' \emph{IEEE Robotics and Automation Letters}, vol.~6, pp. 2791--2798, 2021.

\bibitem{southbay}
W.~Lu, Y.~Zhou, G.~Wan, S.~Hou, and S.~Song, ``{L3-Net}: Towards learning based lidar localization for autonomous driving,'' in \emph{Proc. of the IEEE Conf. on Computer Vision and Pattern Recognition}, Long Beach, CA, June 2019, pp. 6389--6398.

\bibitem{kiss-icp}
I.~Vizzo, T.~Guadagnino, B.~Mersch, L.~Wiesmann, J.~Behley, and C.~Stachniss, ``Kiss-icp: In defense of point-to-point icp--simple, accurate, and robust registration if done the right way,'' \emph{IEEE Robotics and Automation Letters}, vol.~8, pp. 1029--1036, 2023.

\bibitem{ptv3}
X.~Wu, L.~Jiang, P.-S. Wang, Z.~Liu, X.~Liu, Y.~Qiao, W.~Ouyang, T.~He, and H.~Zhao, ``Point transformer v3: Simpler faster stronger,'' in \emph{Proc. of the IEEE/CVF Conf. on Computer Vision and Pattern Recognition}, Seattle, WA, June 2024, pp. 4840--4851.

\end{thebibliography}

\end{document}